\documentclass[nohyperref]{article}

\makeatletter
\newenvironment{chapquote}[2][2em]
  {\setlength{\@tempdima}{#1}%
   \def\chapquote@author{#2}%
   \parshape 1 \@tempdima \dimexpr\textwidth-2\@tempdima\relax%
   \itshape}
  {\par\normalfont\hfill--\ \chapquote@author\hspace*{\@tempdima}\par\bigskip}
\makeatother

\usepackage{microtype}
\usepackage{graphicx}
\usepackage{subfigure}
\usepackage{booktabs} 

\usepackage{xcolor}
\definecolor{citeblue}{RGB}{1,25,108}
\usepackage[pagebackref=false,breaklinks=true,colorlinks=true,citecolor=citeblue,bookmarks=false]{hyperref}
\usepackage{xspace}

\usepackage[accepted]{icml2023}

\usepackage{amsmath,bm}
\usepackage{amssymb}
\usepackage{mathtools}
\usepackage{amsthm}

\usepackage[capitalize,noabbrev]{cleveref}

\theoremstyle{plain}

\theoremstyle{definition}

\theoremstyle{remark}

\usepackage[textsize=tiny]{todonotes}

\definecolor{myMagenta}{rgb}{0.9,0,0.4}

\newcommand{\method}{\textcolor{black}{Composer}\xspace}
\icmltitlerunning{Composer: Creative and Controllable Image Synthesis with Composable Conditions}

\begin{document}

\twocolumn[
\icmltitle{Composer: Creative and Controllable Image Synthesis with \\ Composable Conditions}



\icmlsetsymbol{equal}{*}

\begin{icmlauthorlist}
\icmlauthor{Lianghua Huang}{alibaba}
\icmlauthor{Di Chen}{alibaba}
\icmlauthor{Yu Liu}{alibaba}
\icmlauthor{Yujun Shen}{ant}
\icmlauthor{Deli Zhao}{alibaba}
\icmlauthor{Jingren Zhou}{alibaba}
\end{icmlauthorlist}

\icmlaffiliation{alibaba}{Alibaba Group}
\icmlaffiliation{ant}{Ant Group}

\icmlcorrespondingauthor{Lianghua Huang, Di Chen, Yu Liu}{xuangen.hlh, guangpan.cd, ly103369@alibaba-inc.com}
\icmlcorrespondingauthor{Yujun Shen}{shenyujun0302@gmail.com}
\icmlcorrespondingauthor{Deli Zhao}{zhaodeli@gmail.com}
\icmlcorrespondingauthor{Jingren Zhou}{jingren.zhou@alibaba-inc.com}

\icmlkeywords{Machine Learning, ICML}

\vskip 0.3in
]



\printAffiliationsAndNotice{}  

\begin{abstract}
  %
  Recent large-scale generative models learned on big data are capable of synthesizing incredible images yet suffer from limited controllability.
  This work offers a new generation paradigm that allows flexible control of the output image, such as spatial layout and palette, while maintaining the synthesis quality and model creativity.
  With \textit{compositionality} as the core idea, we first decompose an image into representative factors, and then train a diffusion model with all these factors as the conditions to recompose the input.
  At the inference stage, the rich intermediate representations work as composable elements, leading to a huge design space (\textit{i.e.}, exponentially proportional to the number of decomposed factors) for customizable content creation.
  It is noteworthy that our approach, which we call \textbf{\method}, supports various levels of conditions, such as text description as the \textit{global} information, depth map and sketch as the \textit{local} guidance, color histogram for \textit{low-level} details, \textit{etc.}
  Besides improving controllability,
  we confirm that \method serves as a general framework and 
  facilitates a wide range of classical generative tasks without retraining.
  Code and models will be made available.
\end{abstract}

\section{Introduction}
\label{introduction}
\begin{chapquote}{Noam Chomsky~\cite{Chomsky1965aspectsOT}}
    ``The infinite use of finite means.''
\end{chapquote}
Generative image models conditioned on text can now produce photorealistic and diverse images~\cite{ramesh2022,Saharia2022PhotorealisticTD,Rombach2021HighResolutionIS,Yu2022ScalingAM,Chang2023MuseTG}.
To further achieve customized generation, many recent works extend the text-to-image models by introducing conditions such as segmentation maps~\cite{Rombach2021HighResolutionIS,Wang2022PretrainingIA,Couairon2022DiffEditDS}, scene graphs~\cite{Yang2022DiffusionBasedSG}, sketches~\cite{Voynov2022SketchGuidedTD}, depthmaps~\cite{stablediffusion2}, and inpainting masks~\cite{Xie2022SmartBrushTA,Wang2022ImagenEA}, or by finetuning the pretrained models on a few subject-specific data~\cite{Gal2022AnII,Mokady2022NulltextIF,Ruiz2022DreamBoothFT}.
Nevertheless, these models still provide only a limited degree of controllability for designers when it comes to using them for practical applications.
For example, generative models often struggle to accurately produce images with specifications for semantics, shape, style, and color all at once, which is common in real-world design projects.

We argue that the key to controllable image generation relies not only on conditioning, but even more significantly on \textbf{compositionality}~\cite{Brenden2017Machine}.
The latter can exponentially expand the control space by introducing an enormous number of potential combinations (\textit{e.g.,} a hundred images with eight representations each yield about $\mathbf{100^8}$ combinations).
Similar concepts are explored in
the fields of
language and scene understanding~\cite{Keysers2019MeasuringCG,Johnson2016CLEVRAD}, where the compositionality is termed \textit{compositional generalization}, the skill of recognizing or generating a potentially infinite number of novel combinations from a limited number of known components.

\begin{figure*}[!t]
  \centering
  \includegraphics[width=1.0\linewidth]{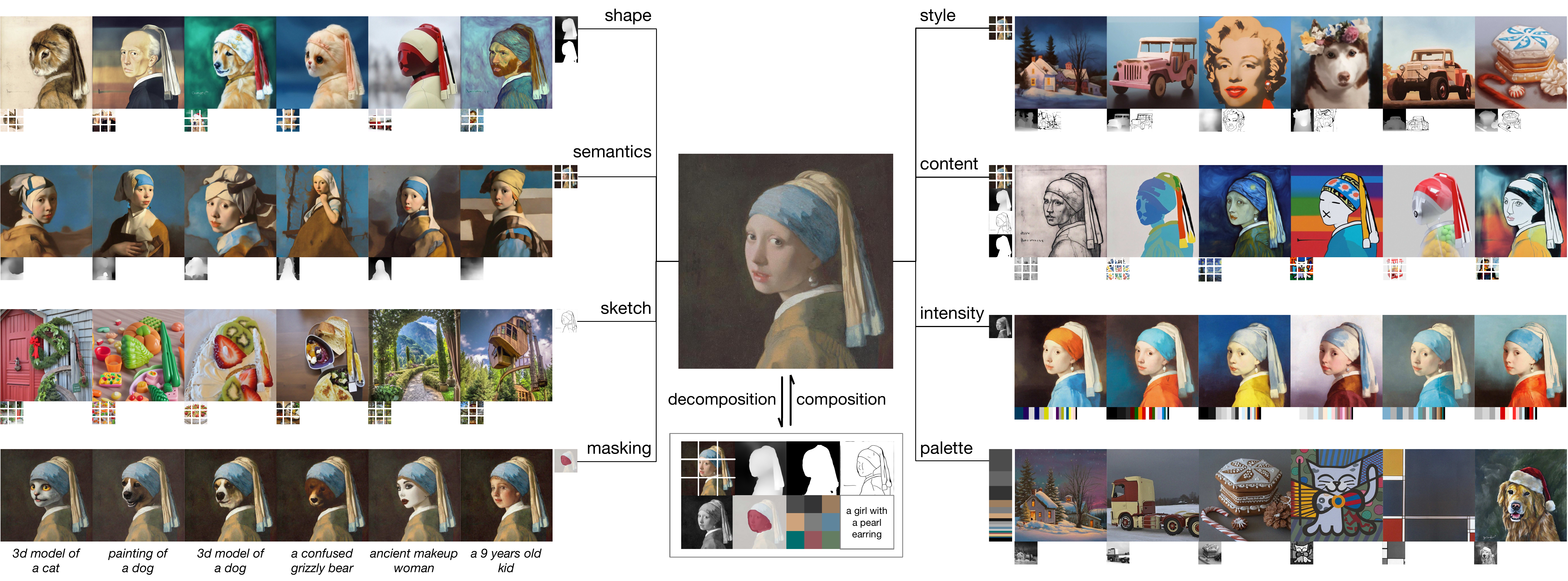}
  \vspace{-20pt}
  \caption{
    \textbf{Concept of compositional image synthesis}, which first decomposes an image to a set of basic components and then recomposes a new one with high creativity and controllability.
    To this end, the components \textit{in various formats} serve as conditions in the generation process and allow \textit{flexible customization} at the inference stage.
    %
    Best viewed in large size.
  }
  \label{fig:figure2}
\end{figure*}

In this work, we build upon the above idea and present Composer, a realization of \textit{compositional generative models}.
By \textit{compositional generative models}, we refer to generative models that are capable of seamlessly recombining visual components to produce new images (\cref{fig:figure2}).
Specifically, we implement Composer as a multi-conditional diffusion model with a UNet backbone~\cite{Nichol2021GLIDETP}.
At every training iteration of Composer, there are two phases:
in the decomposition phase, we break down images in a batch into individual representations using computer vision algorithms or pretrained models;
whereas in the composition phase, we optimize Composer so that it can reconstruct these images from their representation subsets.
Despite being trained with only a reconstruction objective, Composer is capable of decoding novel images from unseen combinations of representations that may come from different sources and potentially incompatible with one another.

While conceptually simple and easy to implement, Composer is surprisingly powerful, enabling encouraging performance on both traditional and previously unexplored image generation and manipulation tasks, including but not limited to: \textit{text-to-image generation, multi-modal conditional image generation, style transfer, pose transfer, image translation, virtual try-on, interpolation and image variation from various directions, image reconfiguration by modifying sketches, depth or segmentation maps, colorization based on optional palettes}, and more.
Moreover, by introducing an orthogonal representation of \textit{masking}, Composer is able to restrict the editable region to a user-specified area for \textit{all} the above operations, more flexible than the traditional inpainting operation, while also preventing modification of pixels outside this region.
Despite being trained in a multi-task manner, Composer achieves a zero-shot FID of 9.2 in text-to-image synthesis on the COCO dataset~\cite{Lin2014MicrosoftCC} when using only caption as the condition, indicating its ability to produce high-quality results.

%

\begin{figure*}[!t]
  \centering
  \includegraphics[width=1.0\linewidth]{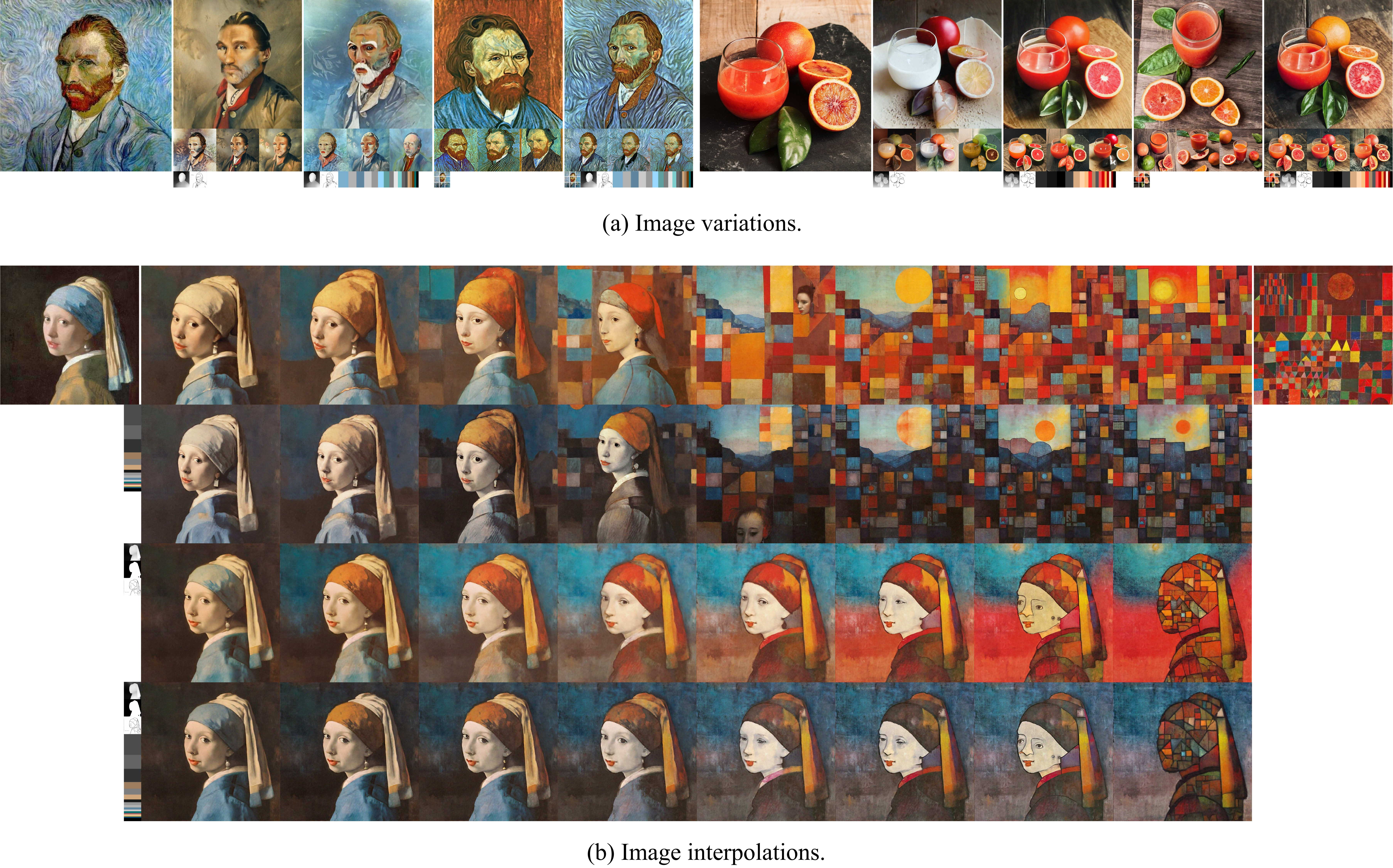}
  \vspace{-7mm}
  \caption{
    (a) \textbf{Image variation.} For each example, the first column shows the source image, while the subsequent four columns are variations of the source image produced by conditioning Composer on different subsets of its representations.
    (b) \textbf{Image interpolation.} On the first row are the results of interpolating all the components between the source image (first column) and the target image (last column). The remaining rows stand for the results where some components (\textit{i.e.}, listed on the left) of the source image are kept unchanged.
  }
  \label{fig:figure3}
\end{figure*}

\section{Method}
Our framework comprises the decomposition phase, where an image is divided into a set of independent components; and the composition phase, where the components are reassembled utilizing a conditional diffusion model.
We first give a brief introduction to diffusion models and the guidance directions enabled by Composer. Subsequently, we explain the implementation of image decomposition and composition in details.

\subsection{Diffusion Models}
\label{sec:diffusion}
Diffusion models~\cite{Ho2020DenoisingDP,Dhariwal2021DiffusionMB,Song2020ImprovedTF,Song2020ScoreBasedGM,Nichol2021GLIDETP} are a type of generative models that produce data from Gaussian noise via an iterative denoising process.
Typically, a simple mean-squared error is used as the denoising objective:
\begin{eqnarray}
  \mathcal{L}_{\text{simple}} = \mathbb{E}_{\mathbf{x}_0, \mathbf{c}, \bm{\epsilon}, t}(\|\bm{\epsilon} - \bm{\epsilon}_{\theta}(a_t \mathbf{x}_0 + \sigma_t \bm{\epsilon}, \mathbf{c})\|_2^2),
\end{eqnarray}
where $\mathbf{x}_0$ are training data with optional conditions $\mathbf{c}$, $t\sim \mathcal{U}(0, 1),~\bm{\epsilon}\sim \mathcal{N}(0, \mathbf{I})$ is the additive Gaussian noise, $a_t,\sigma_t$ are scalar functions of $t$, and $\bm{\epsilon}_{\theta}$ is a diffusion model with learnable parameters $\theta$.
\textit{Classifier-free guidance} is most widely employed in recent works~\cite{Nichol2021GLIDETP,ramesh2022,Rombach2021HighResolutionIS,Saharia2022PhotorealisticTD} for conditional data sampling from a diffusion model, where the predicted noise is adjusted via:
\begin{eqnarray}
  \hat{\bm{\epsilon}}_{\theta}(\mathbf{x}_t, \mathbf{c}) = \omega \bm{\epsilon}_{\theta}(\mathbf{x}_t, \mathbf{c}) + (1 - \omega) \bm{\epsilon}_{\theta}(\mathbf{x}_t),
\end{eqnarray}
where $\mathbf{x}_t = a_t \mathbf{x}_0 + \sigma_t \bm{\epsilon}$, and $\omega$ is a guidance weight.
Sampling algorithms such as DDIM~\cite{Song2020DenoisingDI} and DPM-Solver~\cite{Lu2022DPMSolverAF,Lu2022DPMSolverFS,Bao2022AnalyticDPMAA} are often adopted to speed up the sampling process of diffusion models.
DDIM can also be utilized to deterministically reverse a sample $\mathbf{x}_0$ back to its pure noise latent $\mathbf{x}_T$, enabling various image editing operations.

\textit{Guidance directions:} 
Composer is a diffusion model accepting multiple conditions, which enables various directions
with
classifier-free guidance:
\begin{eqnarray}
  \hat{\bm{\epsilon}}_{\theta}(\mathbf{x}_t, \mathbf{c}) = \omega \bm{\epsilon}_{\theta}(\mathbf{x}_t, \mathbf{c}_2) + (1 - \omega) \bm{\epsilon}_{\theta}(\mathbf{x}_t, \mathbf{c}_1),
\end{eqnarray}
where $\mathbf{c}_1$ and $\mathbf{c}_2$ are two sets of conditions.
Different choices of $\mathbf{c}_1$ and $\mathbf{c}_2$ represent different emphasis on conditions.
Conditions within $(\mathbf{c}_2 \setminus \mathbf{c}_1)$ are emphasized with a guidance weight of $\omega$, those within $(\mathbf{c}_1 \setminus \mathbf{c}_2)$ are suppressed with a guidance weight of $(1 - \omega)$, and conditions within $\mathbf{c}_1\cap \mathbf{c}_2$ are given a guidance weight of $1.0$.

\textit{Bidirectional guidance:}
By reversing an image $\mathbf{x}_0$ to its latent $\mathbf{x}_T$ using condition $\mathbf{c}_1$, and then sampling  from $\mathbf{x}_T$ using another condition $\mathbf{c}_2$, we are able to manipulate the image in a disentangled manner using Composer, where the manipulation direction is defined by the difference between $\mathbf{c}_2$ and $\mathbf{c}_1$.
Similar scheme is also used in~\cite{Wallace2022EDICTED}.
We use this approach in \cref{sec:manipulations} and \cref{sec:reformulations}.

\subsection{Decomposition}
We decompose an image into decoupled representations which capture various aspects of it.
We describe eight representations we use in this work, where all of them are extracted on-the-fly during training.

\textit{Caption:}
We directly use title or description information in image-text training data (\textit{e.g.,} LAION-5B~\cite{Schuhmann2022LAION5BAO}) as image captions.
One can also leverage pretrained image captioning models when annotations are not available.
We represent these captions using their sentence and word embeddings extracted by the pretrained CLIP ViT-L/14@336px~\cite{radford21} model.

\begin{figure*}[!t]
  \centering
  \includegraphics[width=1.0\linewidth]{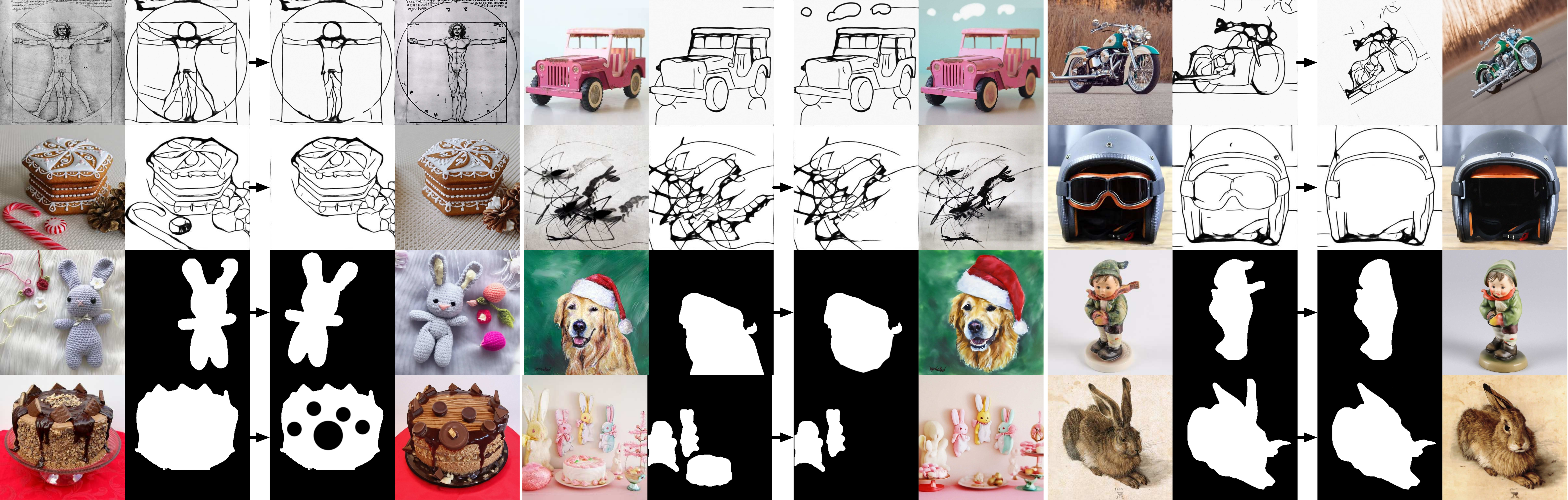}
  \vspace{-5mm}
  \caption{
    \textbf{Image reconfiguration.}
    Composer supports reconfiguring an image simply by altering its representations, such as sketch and segmentation map.
  }
  \label{fig:figure4}
\end{figure*}

\textit{Semantics and style:}
We use the image embedding extracted by the pretrained CLIP ViT-L/14@336px~\cite{radford21} model to represent the semantics and style of an image, similar to unCLIP~\cite{ramesh2022}.

\textit{Color:}
We represent the color statistics of an image using the smoothed CIELab histogram~\cite{sergeyk16}.
We quantize the CIELab color space to 11 hue values, 5 saturation values, and 5 light values, and we use a smoothing sigma of 10. We empirically find these settings to work well.

\textit{Sketch:}
We apply an edge detection model~\cite{su21} followed by a sketch simplification algorithm~\cite{SimoSerra17} to extract the sketch of an image.
Sketches capture local details of images and have less semantics.

\textit{Instances:}
We apply instance segmentation on an image using the pretrained YOLOv5~\cite{Jocher20} model to extract its instance masks.
Instance segmentation masks reflect the category and shape information of visual objects.

\textit{Depthmap:}
We use a pretrained monocular depth estimation model~\cite{Ranftl2022} to extract the depthmap of an image, which roughly captures the image's layout.

\textit{Intensity:}
We introduce raw grayscale images as a representation to force the model to learn a disentangled degree of freedom for manipulating colors.
To introduce randomness, we uniformly sample from a set of predefined RGB channel weights to create grayscale images.

\textit{Masking:}
We introduce image masks to enable Composer to restrict image generation or manipulation to an editable region.
We use a 4-channel representation, where the first 3 channels correspond to the masked RGB image, while the last channel corresponds to the binary mask.

It should be noted that, while this work experiments with eight conditions described above, users are allowed to freely customize their conditions using Composer.


\begin{figure*}[!t]
  \centering
  \includegraphics[width=1.0\linewidth]{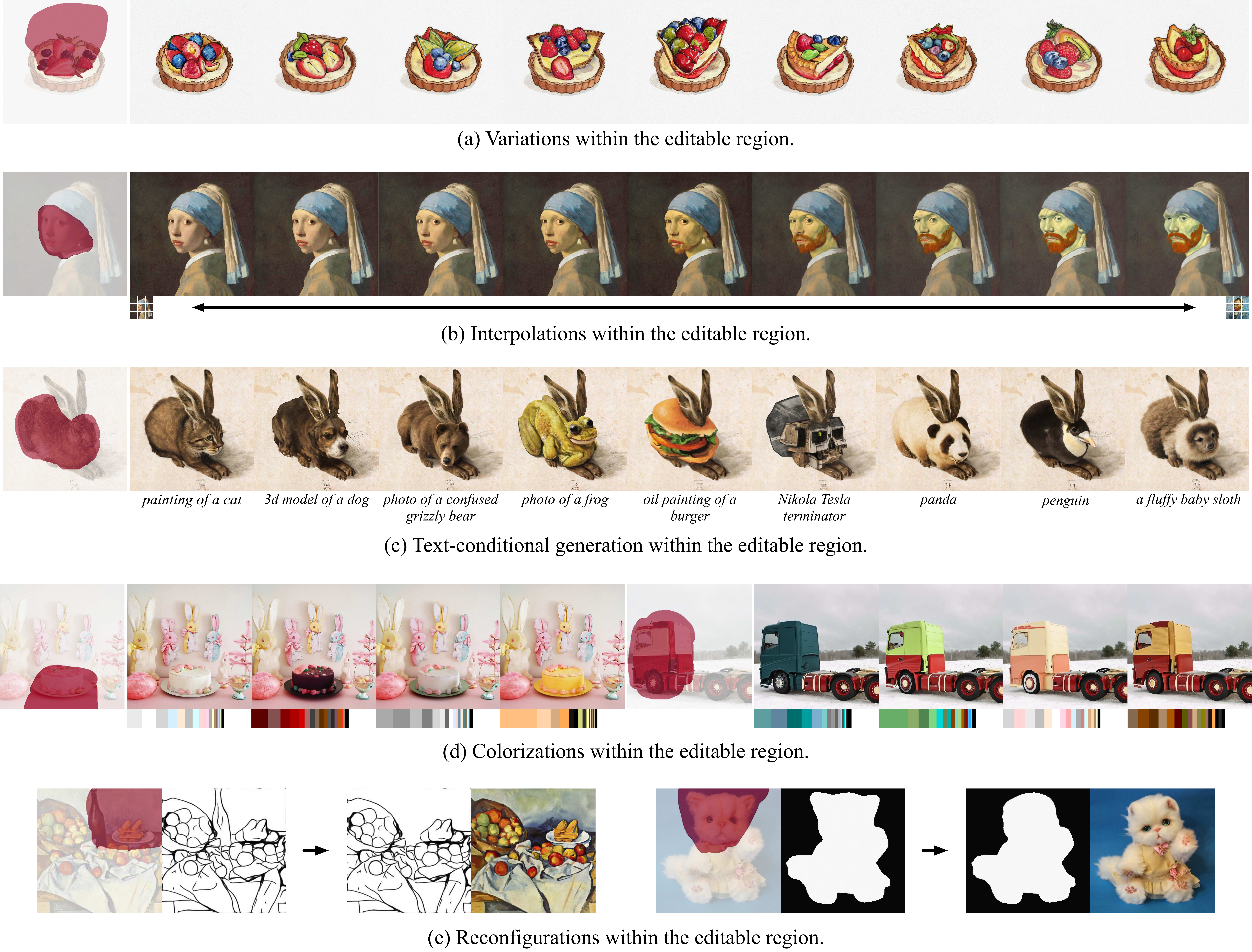}
  \vspace{-5mm}
  \caption{
    \textbf{Region-specific image editing.}
    Through introducing a masked image as an additional condition, Composer manages to direct the manipulation to the region of interest. 
  }
  \label{fig:figure5}
\end{figure*}

\subsection{Composition}
\label{sec:composition}
We use diffusion models to \textit{recompose} images from a set of representations.
Specifically, we leverage the GLIDE~\cite{Nichol2021GLIDETP} architecture and modify its conditioning modules.
We explore two different mechanisms to condition the model on our representations:

\textit{Global conditioning:}
For global representations including CLIP sentence embeddings, image embeddings and color palettes, we project and add them to the timestep embedding.
In addition, we project image embeddings and color palettes into eight extra tokens and concatenate them with CLIP word embeddings, which are then used as the context for cross-attention in GLIDE, similar to unCLIP~\cite{ramesh2022}.
Since conditions are either additive or can be selectively masked in cross-attention, it is straightforward to either drop conditions during training and inference, or to introduce new global conditions.

\textit{Localized conditioning:}
For localized representations including sketches, segmentation masks, depthmaps, intensity images, and masked images, we project them into uniform-dimensional embeddings with the same spatial size as the noisy latent $\mathbf{x}_t$
using stacked convolutional layers.
We then compute the sum of these embeddings and concatenate the result to $\mathbf{x}_t$ before feeding it into the UNet.
Since the embeddings are additive, it is easy to accommodate for missing conditions or to incorporate new localized conditions.


\textit{Joint training strategy:}
It is essential to devise a joint training strategy that enables the model to learn to decode images from a variety of combinations of conditions.
We experiment with several configurations and identify a simple yet effective configuration, where we use an independent dropout probability of 0.5 for each condition, a probability of 0.1 for dropping all conditions, and a probability of 0.1 for retaining all conditions.
We use a special dropout probability of 0.7 for intensity images because they contain the vast majority of information about the images and may underweight other conditions during training.

The base diffusion model produces images of $64\times 64$ resolution.
To generate high-resolution images, we train two unconditional diffusion models for upsampling to respectively upscale images from $64\times 64$ to $256\times 256$ and from $256\times 256$ to $1024\times 1024$ resolutions.
The architectures of the upsampling models are modified from unCLIP~\cite{ramesh2022}, where we use more channels in low-resolution layers and introduce self-attention blocks to scale up the capacity.
We also introduce an optional prior model~\cite{ramesh2022} that produces image embeddings from captions. We empirically find that the prior model is capable of improving the diversity of generated images for certain combinations of conditions.

\begin{figure*}[!t]
  \centering
  \includegraphics[width=1.0\linewidth]{figs_compressed/figure5}
  \vspace{-5mm}
  \caption{
    \textbf{Reformulation of traditional image generation tasks} using our Composer.
    Note that the model is directly applied to all tasks without any retraining, highlighting the potential and flexibility of the proposed compositional generation framework.
  }
  \label{fig:figure6}
\end{figure*}

\section{Experiments}

\subsection{Training Details}

We train a 2B parameter base model for conditional image generation at $64\times 64$ resolution, a 1.1B parameter model for upscaling images to $256\times 256$ resolution, and a 300M parameter model for further upscaling images to $1024\times 1024$ resolution. Additionally, we trained a 1B parameter prior model for optionally projecting captions to image embeddings. We use batch sizes of 4096, 1024, 512, and 512 for the prior, base, and two upsampling models, respectively.
We train on a combination of public datasets, including ImageNet21K~\cite{Russakovsky2014ImageNetLS}, WebVision~\cite{Li2017WebVisionDV}, and a filtered version of the LAION dataset~\cite{Schuhmann2022LAION5BAO} with around 1B images.
We eliminate duplicates, low resolution images, and images potentially contain harmful content from the LAION dataset.
For the base model, we pretrain it with 1M steps on the full dataset using only image embeddings as the condition, and then finetune the model on a subset of 60M examples (excluding LAION images with aesthetic scores below 7.0) from the original dataset for 200K steps with all conditions enabled.
The prior and upsampling models are trained for 1M steps on the full dataset.


\begin{figure*}[!t]
  \centering
  \includegraphics[width=0.97\linewidth]{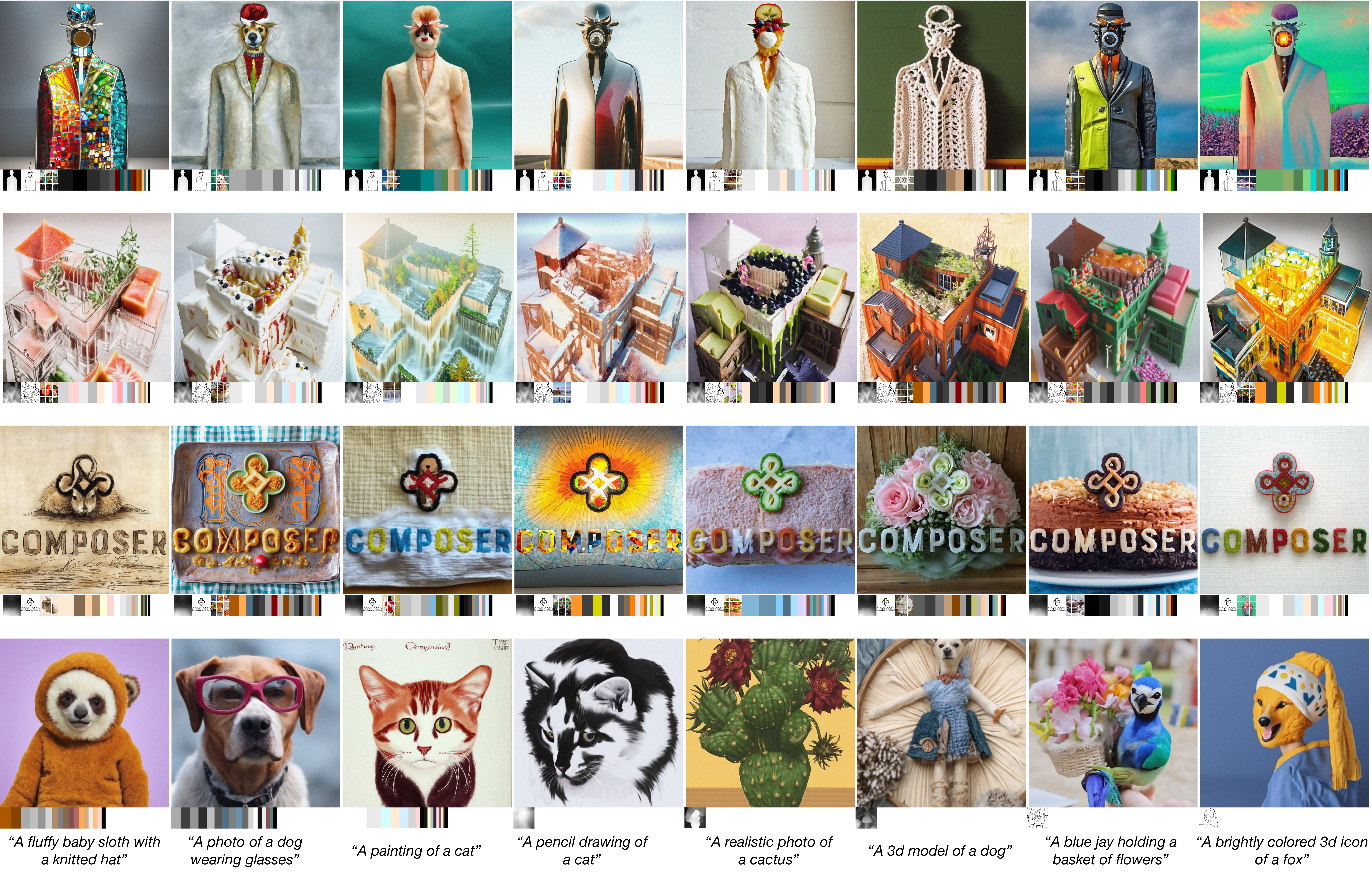}
  \vspace{-2mm}
  \caption{
    \textbf{Compositional image generation} results produced by Composer.
    The conditions used to generate each image are presented below the image.
    Best viewed in large size. 
  }
  \label{fig:figure1}
\end{figure*}

\subsection{Image Manipulation}
\label{sec:manipulations}

\textit{Variations:}
Using Composer, we can create new images that are similar to a given image but vary in certain aspects by conditioning on a specific subset of its representations.
By carefully selecting combinations of different representations, we have a great degree of flexibility to control the scope of image variations (\cref{fig:figure3}a).
When more conditions are incorporated, our approach easily yields more accurate reconstructions than unCLIP~\cite{ramesh2022}, which is conditioned solely on image embeddings.

\textit{Interpolations:}
By traversing in the embedding space of global representations between two images, we can blend the two images for variations.
Composer further gives us precise control over which elements to interpolate between two images and which to keep unchanged, resulting in a multitude of interpolation directions (\cref{fig:figure3}b).

\textit{Reconfigurations:}
Image reconfiguration~\cite{Sun2019ImageSF} refers to manipulating an image through direct modification of one or more of its representations.
Composer offers a variety of options for image reconfiguration (\cref{sec:diffusion}).
Specifically, given an image $\mathbf{x}$, we can obtain its latent $\mathbf{x}_T$ by applying DDIM inversion conditioned on a set of its representations $\mathbf{c}_i$; we then apply DDIM sampling starting from $\mathbf{x}_T$ conditioned on a modified set of representations $\mathbf{c}_j$ to obtain a variant of the image $\hat{\mathbf{x}}$.
The variant $\hat{\mathbf{x}}$ is expected to differ from $\mathbf{x}$ along the variation direction defined by the difference between $\mathbf{c}_j$ and $\mathbf{c}_i$, but they are otherwise similar.
By following this process, we are able to manipulate an image from diverse directions (\cref{fig:figure4}).

\textit{Editable region:}
By conditioning Composer on a set of representations $\textbf{c}$ along with a masked image $\textbf{m}$, it is possible to restrict the variations within the area defined by $\textbf{m}$.
Remarkably, editable region is orthogonal to all image generation and manipulation operations, offering Composer substantially greater flexibility of image editing than mere inpainting (\cref{fig:figure5}).

\subsection{Reformulation of Traditional Generation Tasks}
\label{sec:reformulations}

Many traditional image generation and manipulation tasks can be reformulated using the Composer architecture.
Below we describe several examples.

\textit{Palette-based colorization:}
There are two methods to colorize an image $\mathbf{x}$ according to palette $\mathbf{p}$ using Composer: one entails conditioning the sampling process on both the grayscale version of $\mathbf{x}$ and $\mathbf{p}$, while the other involves applying a \textit{reconfiguration} (\cref{sec:diffusion}) on $\mathbf{x}$ in terms of color palette.
We find the latter approach yields more reasonable and diverse results and we use it in \cref{fig:figure6}a.

\textit{Style transfer:}
Composer roughly disentangles the content and style representations, which allows us to transfer the style of image $\mathbf{x}_1$ to another image $\mathbf{x}_2$ by simply conditioning on the style representations of $\mathbf{x}_1$ and the content representations of $\mathbf{x}_2$.
It is also possible to control the transfer strength by interpolating style representations between the two images.
We show examples in \cref{fig:figure6}b.

\textit{Image translation:}
Image translation refers to the task of transforming an image to a variant with content kept unchanged but style converted to match a target domain.
We use all available representations of an image to depict its content, with a text description to capture the target domain.
We leverage the reconfiguration approach described in \cref{sec:diffusion} to manipulate images (\cref{fig:figure6}c).

\textit{Pose transfer:}
The CLIP embedding of an image captures its style and semantics, enabling Composer to modify the pose of an object without compromising its identity.
We use the object's segmentation map to represent its pose and the image embedding to capture its semantics, then leverage the reconfiguration approach described in \cref{sec:diffusion} to modify the pose of the object (\cref{fig:figure6}d).

\textit{Virtual try-on:}
Given a garment image $\mathbf{x}_1$ and a body image $\mathbf{x}_2$, we can first mask the clothes in $\mathbf{x}_2$, and then condition the sampling process on the masked image ${\mathbf{m}}_2$ along with the CLIP image embedding of $\mathbf{x}_1$ to produce a virtual try-on result (\cref{fig:figure6}e).
Despite moderate quality, 
the results demonstrate the possibilities of Composer to cope with difficult problems with one unified framework.


\subsection{Compositional Image Generation}
By conditioning Composer on a combination of visual components from different sources, it is possible to produce an enormous number of generation results from a limited set of materials.
\cref{fig:figure1} shows some selected examples.

\subsection{Text-to-Image Generation}
To further assess Composer's image generation quality, we compare its performance with the state-of-the-art text-to-image generation models on the COCO dataset~\cite{Lin2014MicrosoftCC}. We use sampling steps of 100, 50, and 20 for the prior, base, and $64\times 64$ to $256\times 256$ upsampling models respectively and a guidance scale of $3.0$ for the prior and base models. Despite its multi-task training, Composer achieves an competitive FID score of 9.2 and a CLIP score of 0.28 on COCO, comparable to the best-performing models.


\section{Related Work}
Diffusion models~\cite{Ho2020DenoisingDP,Nichol2021ImprovedDD,Dhariwal2021DiffusionMB,Rombach2021HighResolutionIS,Nichol2021GLIDETP,ramesh2022,stablediffusion2,Saharia2022PhotorealisticTD} are emerging as a successful paradigm for image generation, outperforming GANs~\cite{Xu2017AttnGANFT,Zhu2019DMGANDM,Zhang2021CrossModalCL} and comparable to autoregressive models~\cite{Ramesh2021ZeroShotTG,Yu2021VectorquantizedIM,Esser2020TamingTF,Yu2022ScalingAM,Ding2021CogViewMT} in terms of fidelity and diversity.
Our method builds on recent hierarchical diffusion models~\cite{ramesh2022,Saharia2022PhotorealisticTD}, where one large diffusion model is used to produce small-resolution images, while two relatively smaller diffusion models upscale the images to higher resolutions.
However, unlike these text-to-image models, our method supports composable conditions and exhibits better flexibility and controllability.

Many recent works extend pretrained text-to-image diffusion models to achieve multi-modal or customized generation, typically by introducing conditions such as inpainting masks~\cite{Xie2022SmartBrushTA,Wang2022ImagenEA}, sketches~\cite{Voynov2022SketchGuidedTD}, scene graphs~\cite{Yang2022DiffusionBasedSG}, keypoints~\cite{Li2023GLIGENOG}, segmentation maps~\cite{Rombach2021HighResolutionIS,Wang2022PretrainingIA,Couairon2022DiffEditDS}, a composition of multiple text descriptions~\cite{Liu2022CompositionalVG}, and depthmaps~\cite{stablediffusion2}, or by finetuning parameters on a few subject-specific data~\cite{Gal2022AnII,Mokady2022NulltextIF,Ruiz2022DreamBoothFT}.
Besides, GAN-based methods can also accept a combination of multiple conditions to enable controllable generation~\cite{Huang2021MultimodalCI}.
Compared to these approaches, Composer merits the compositionality across conditions, enabling a larger control space and greater flexibility in image generation and manipulations.

\section{Conclusion and Discussion}
Our decomposition-composition paradigm demonstrates that when conditions are composable rather than used independently, the control space of generative models can be vastly expanded.
As a result, a broad range of traditional generative tasks can be reformulated using our Composer architecture, and previously unexplored generative abilities are unveiled, motivating further research into a variety of decomposition algorithms that can achieve increased controllability.
In addition, we present multiple ways to utilize Composer for a range of image generation and manipulation tasks based on classifier-free and bidirectional guidance, giving useful references for future research.

Although we empirically find a simple, workable configuration for joint training of multiple conditions in \cref{sec:composition}, the strategy is not perfect, \textit{e.g.,} it may downweight the single-conditional generation performance.
For example, without access to global embeddings, sketch- or depth-based generation usually produces relatively dark images.
Another issue is that conflicts may exist when incompatible conditions occur. For instance, text embeddings are often downweighted in generated results when image and text embeddings with different semantics are jointly used.

Previous studies~\cite{Nichol2021GLIDETP,ramesh2022,Saharia2022PhotorealisticTD} highlight the potential risks associated with image generation models, such as deceptive and harmful content.
Composer's improvements in controllability further raise this risk.
We intend to thoroughly investigate how Composer can mitigate the risk of misuse
and possibly creating a filtered version before making the work public.


\bibliography{main_v2}
\bibliographystyle{icml2023}

\newpage
\appendix
\onecolumn

\section{Architecture Details}
\begin{table}[h]
    \setlength\tabcolsep{4pt}
    \begin{center}
    \begin{small}
    \begin{tabular}{lccccc}
    \toprule
                         & Prior & $64$ & $64 \rightarrow 256$ & $256 \rightarrow 1024$\\
    \midrule
    Diffusion steps      & 1000    & 1000    & 1000    & 1000 \\
    Noise schedule       & cosine  & cosine  & cosine  & linear \\
    Sampling steps       & 100      & 50     & 20      & 10 \\
    Sampling variance method & dpm-solver & dpm-solver & dpm-solver    & dpm-solver \\
    Model size           & 1B      & 2B    & 1.1B    & 300M \\
    Channels             & -       & 512     & 320     & 192 \\
    Depth                & -       & 3       & 3       & 2 \\
    Channels multiple    & -       & 1,2,3,4 & 1,2,3,5 & 1,1,2,2,4,4 \\
    Heads channels       & -       & 64      & 64       & - \\
    Attention resolution & -       & 32,16,8 & 32,16       & - \\
    Dropout              & -       & 0.1     & 0.1     & - \\
    Weight decay         & 6.0e-2  & -       & -       & - \\
    Batch size           & 4096    & 1024    & 512    & 512 \\
    Iterations           & 1M    & 1M    & 1M      & 1M \\
    Learning rate        & 1.1e-4  & 1.2e-4  & 1.1e-4  & 1.0e-4 \\
    Adam $\beta_2$       & 0.96    & 0.999   & 0.999   & 0.999 \\
    Adam $\epsilon$      & 1.0e-6   & 1.0e-8    & 1.0e-8    & 1.0e-8  \\
    EMA decay            & 0.9999  & 0.9999  & 0.9999  & 0.9999 \\
    \bottomrule
    \end{tabular}
    \end{small}
    \end{center}
    \caption{Hyperparameters for Composer. We use DPM-Solver++~\cite{Lu2022DPMSolverFS} as the sampling algorithm for all diffusion models.}
    \label{tab:hps}
    \vskip -0.2in
\end{table}

\newpage
\section{Conditioning Modules}
\begin{figure}[th]
  \centering
  \includegraphics[width=0.26\linewidth]{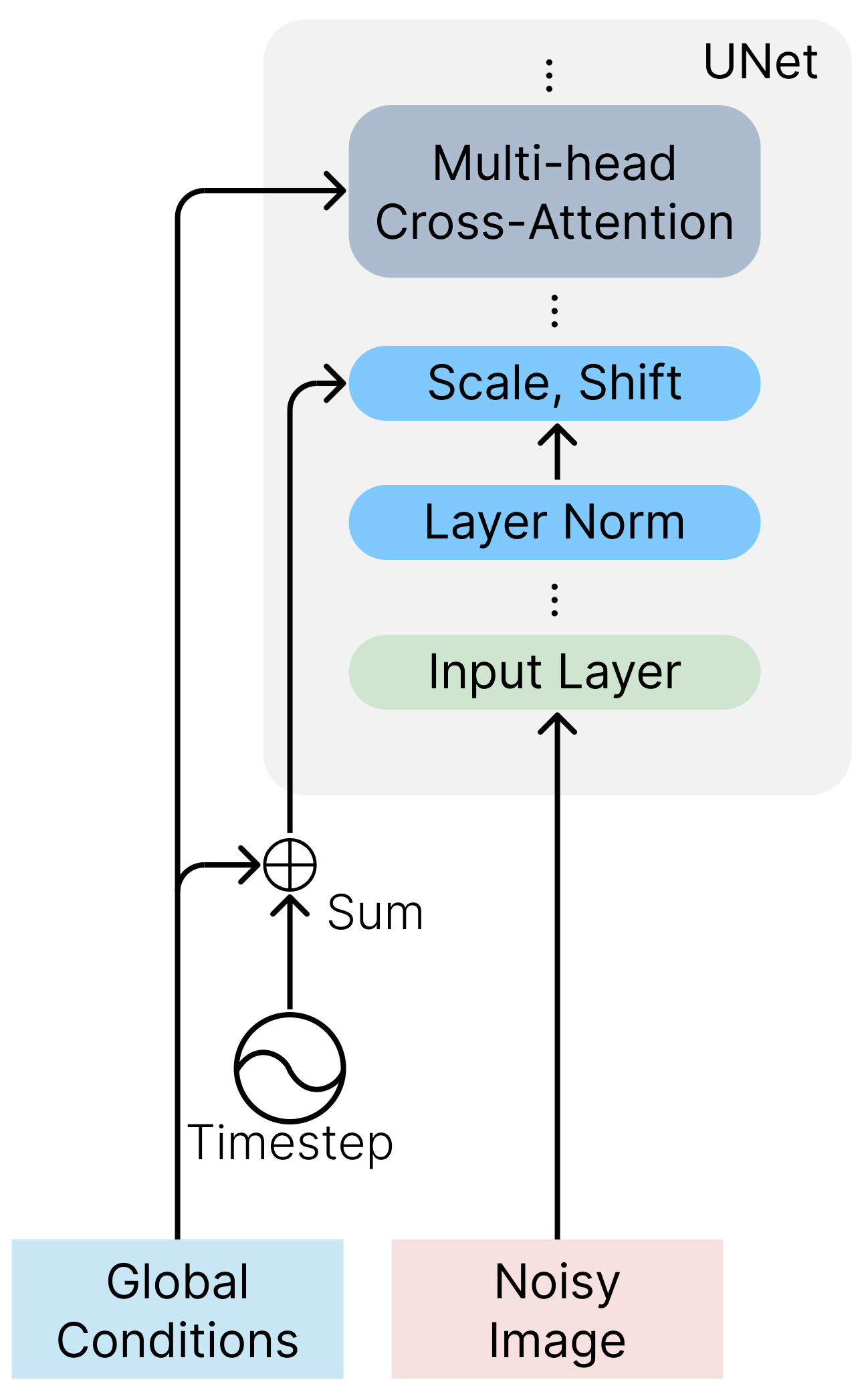}
  \vspace{-3mm}
  \caption{
    Global conditioning module of Composer.
    For global conditions such as CLIP sentence embeddings, image embeddings, and color histograms, we project and add them to the timestep embedding.
    Moreover, we project image embeddings and color palettes into eight extra tokens and concatenate them with CLIP word embeddings, which are then used as the context input for cross-attention layers.
  }
  \label{fig:global}
\end{figure}
\begin{figure}[th]
  \centering
  \includegraphics[width=0.26\linewidth]{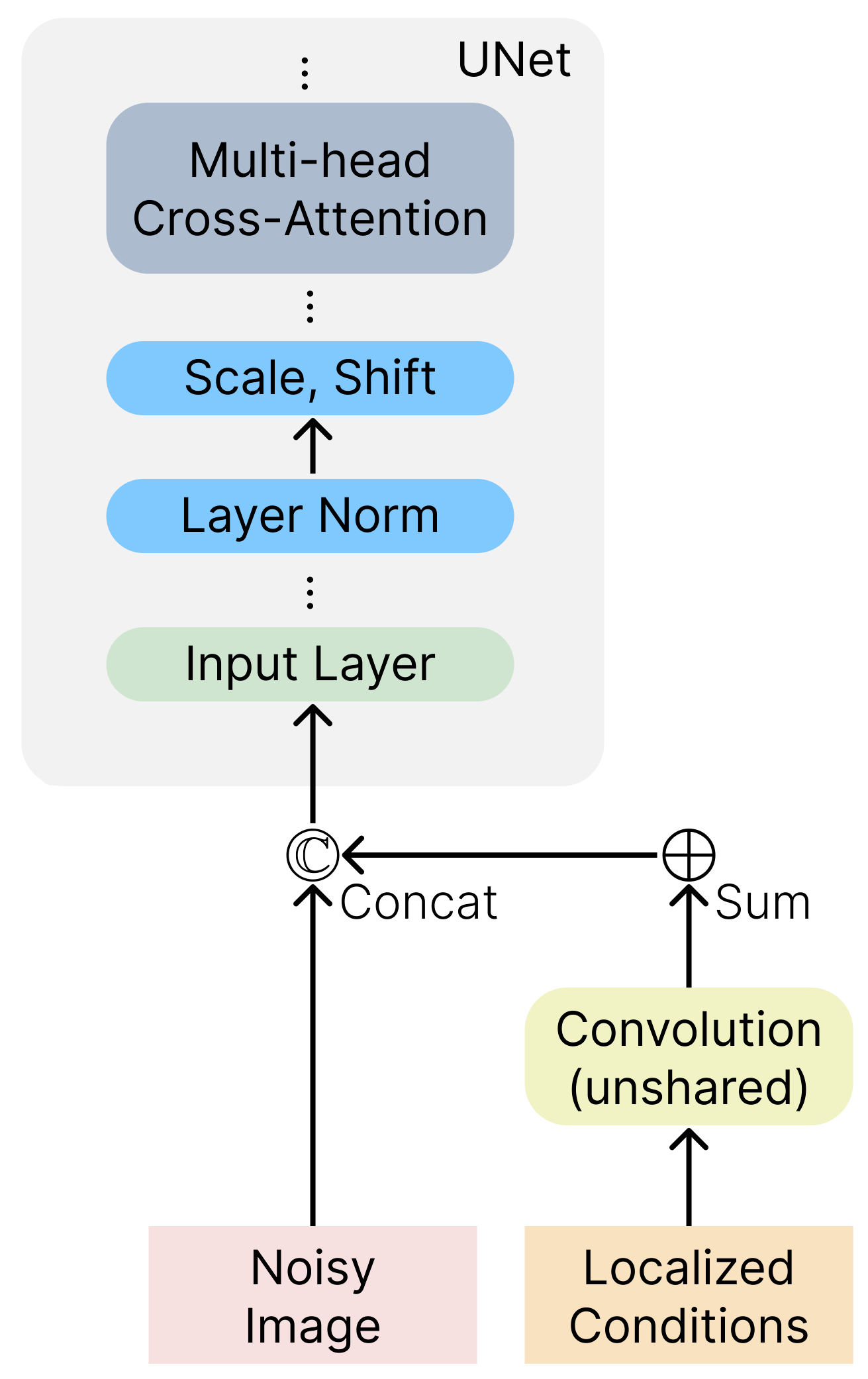}
  \vspace{-3mm}
  \caption{
    Local conditioning module of Composer. For local conditions such as segmentation maps, depthmaps, sketches, grayscale images, and masked images, we project them into uniform-dimensional embeddings with the same spatial size as the noisy image using stacked convolutional layers.
    Subsequently, we compute the sum of these embeddings and concatenate the result to the noisy image.
  }
  \label{fig:local}
\end{figure}

\newpage

\section{Additional Samples}

\begin{figure*}[bh]
  \centering
  \includegraphics[width=1.0\linewidth]{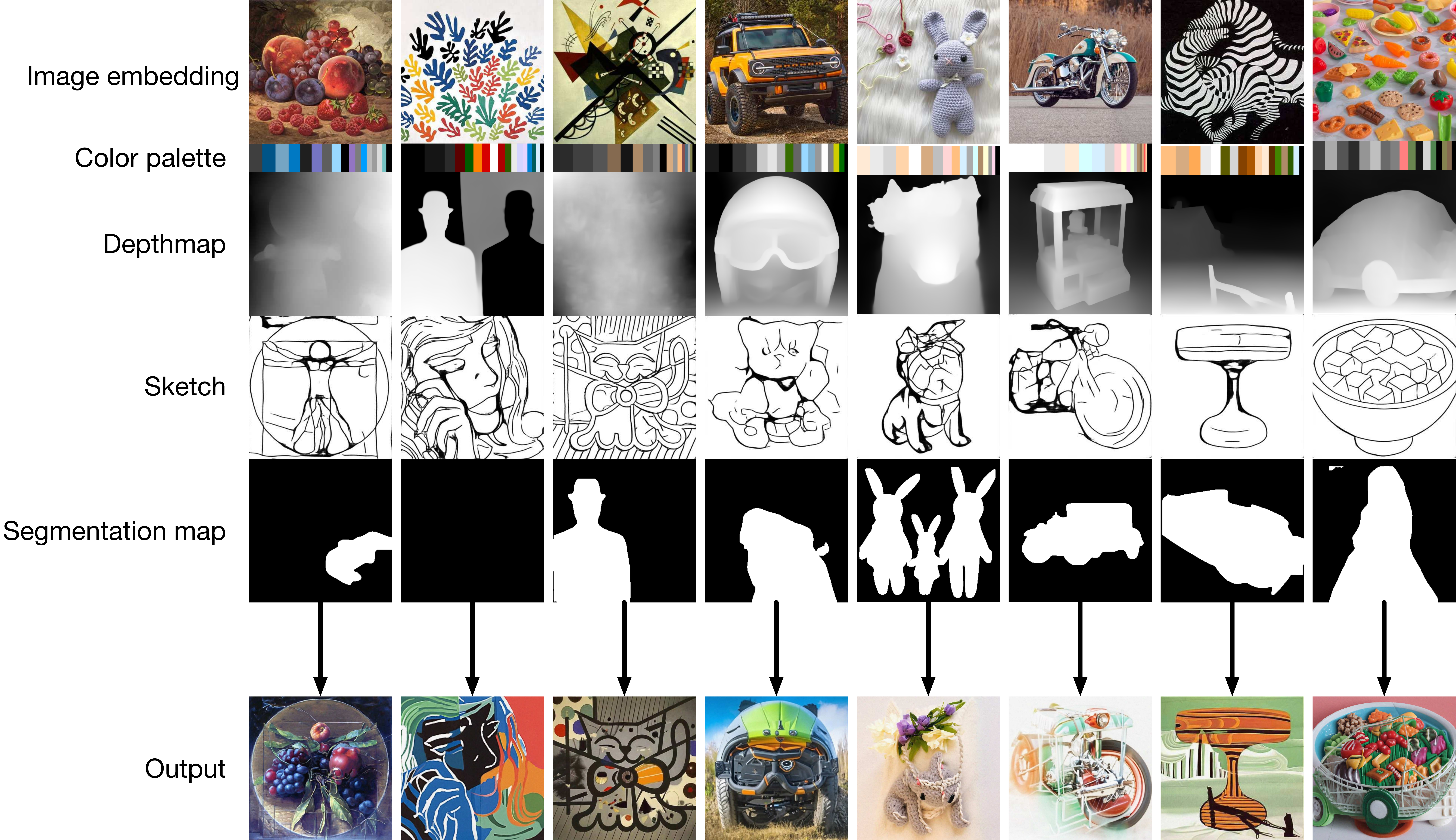}
  \vspace{-3mm}
  \caption{
    This figure indicates how Composer resolves conflicting conditions by illustrating extreme cases in which the conditions come from disparate sources.
    One of our observations is that Composer typically gives less weight to conditions with fewer details when conflicts exist, such as segmentation maps in comparison to sketches and depthmaps, and text embeddings versus image embeddings.
  }
  \label{fig:conflicts}
\end{figure*}

\begin{figure*}[!t]
  \centering
  \includegraphics[width=1.0\linewidth]{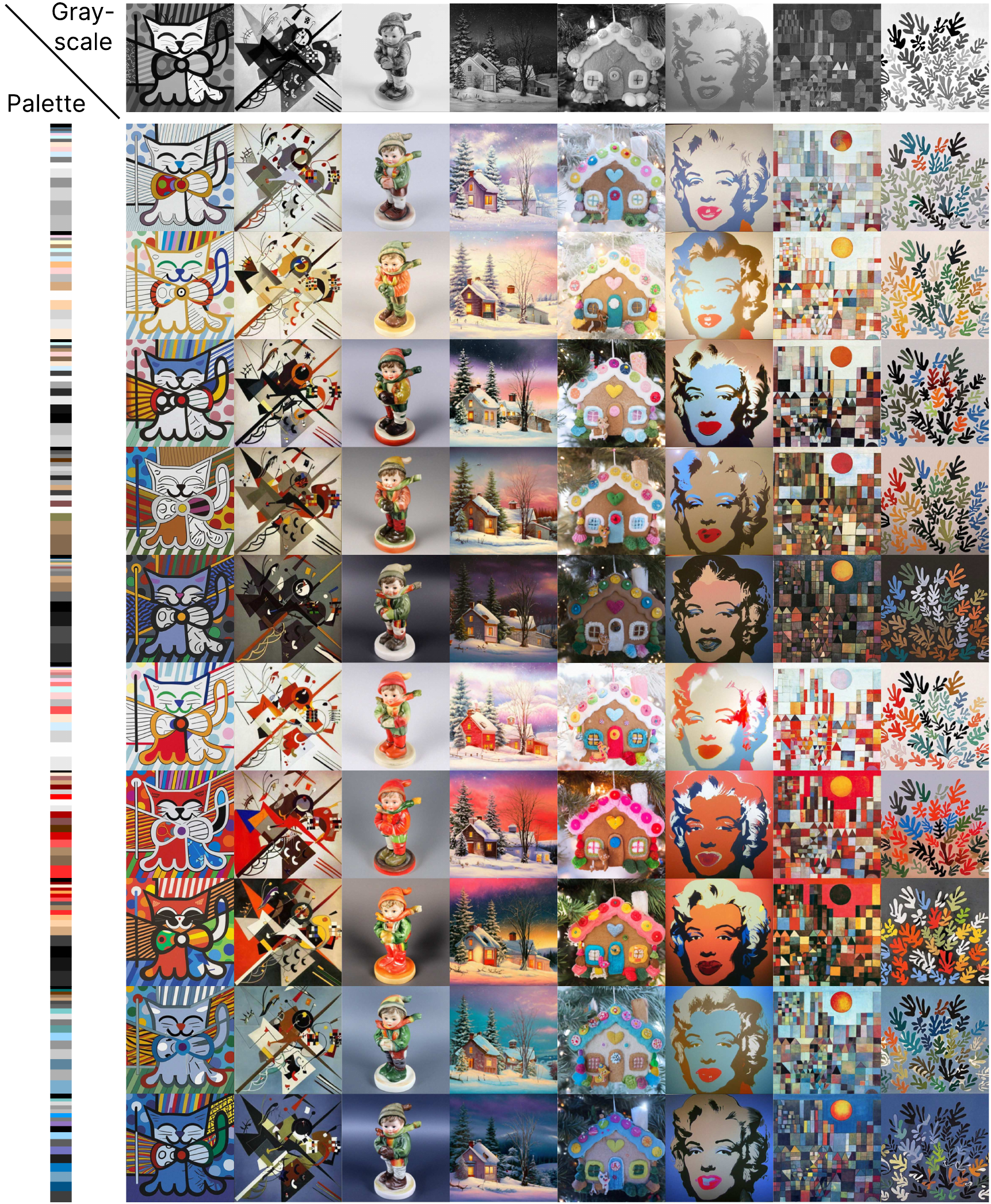}
  \vspace{-3mm}
  \caption{
    Additional colorization results, visualized at a resolution of $256\times 256$ to reduce file size.
  }
  \label{fig:colorization}
\end{figure*}

\begin{figure*}[!t]
  \centering
  \includegraphics[width=1.0\linewidth]{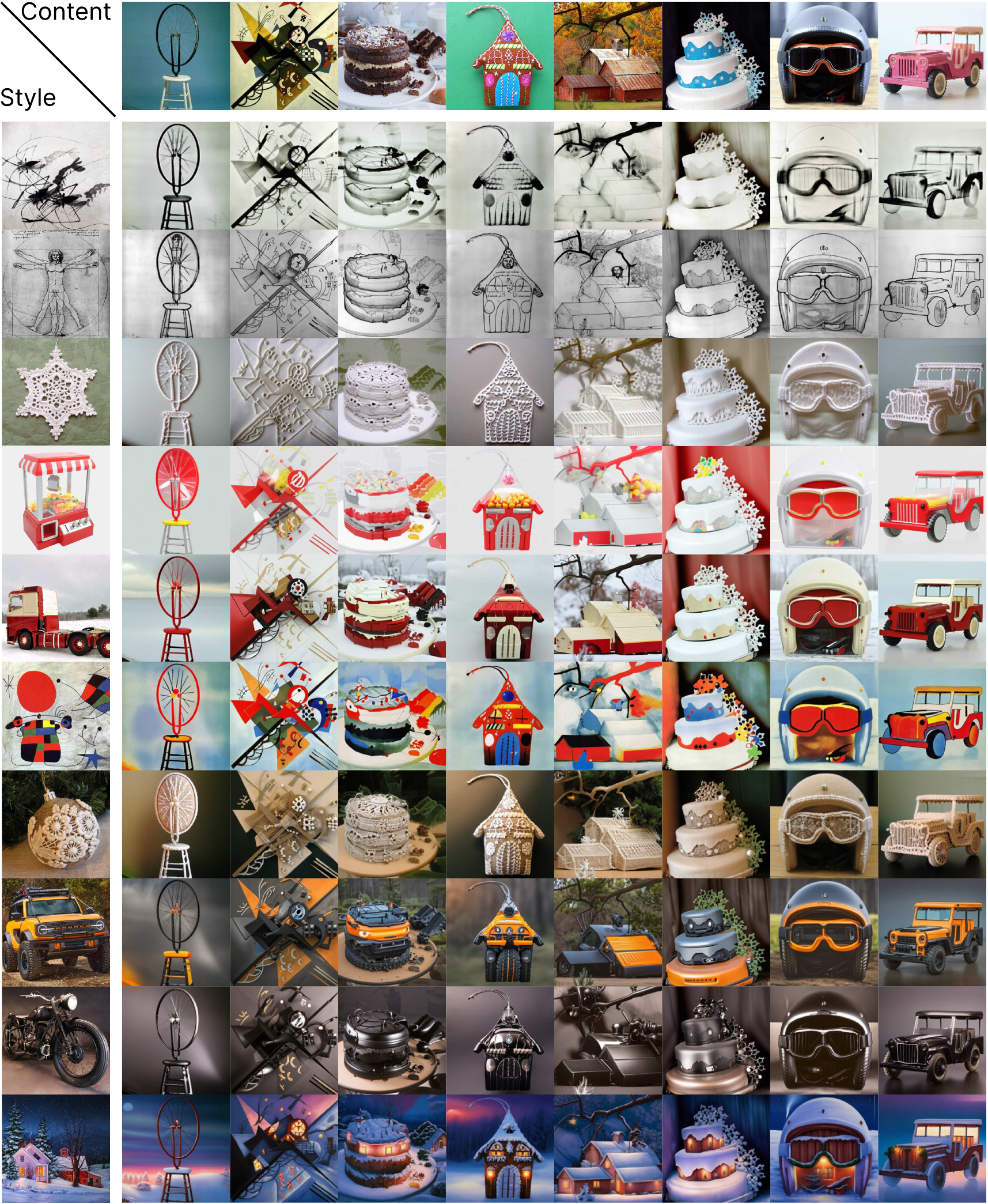}
  \vspace{-3mm}
  \caption{
    Additional style transfer results, visualized at a resolution of $256\times 256$ to reduce file size.
  }
  \label{fig:style_transfer}
\end{figure*}


\end{document}